%% file: template.tex
\documentclass{article}
\usepackage{spconf,amsmath,graphicx,algorithm,algorithmic,booktabs,multirow, multicol,float,amsfonts}
\usepackage{enumitem}
\usepackage{subfig}
\usepackage{amsthm}
\usepackage{amsfonts}
\usepackage{stackengine}
\newtheorem{prop}{Proposition}
\newtheorem{lemma}{Lemma}
\usepackage{cite}
\usepackage{hyperref}
\usepackage{appendix}

\title{Interpretation of Neural Networks is Susceptible to \\ Universal Adversarial Perturbations}
\twoauthors
 {Haniyeh Ehsani Oskouie}{Sharif University of Technology\\
	Department of Computer Engineering\\
	haniyeh.ehsani@sharif.edu}
	{Farzan Farnia}{
 The Chinese University of Hong Kong\\
	Department of Computer Science \& Engineering\\
    farnia@cse.cuhk.edu.hk}
	
\begin{document}
\ninept
%
\maketitle
\begin{abstract}
\input{0-abstract.tex}
\end{abstract}
\section{Introduction}
\input{1-introduction.tex}

\section{Preliminaries}
\input{2-prelim.tex}

\section{Universal Perturbations for Interpretation}

\input{3-algorithm.tex}

\section{Numerical Results}

\input{4-numericals.tex}


{
\bibliographystyle{IEEEbib}
\bibliography{ref}}

\clearpage
\begin{appendices}
\input{5-appendix.tex}
\end{appendices}

\end{document}

%% file: 0-abstract.tex
Interpreting neural network classifiers using gradient-based saliency maps has been extensively studied in the deep learning literature. While the existing algorithms manage to achieve satisfactory performance in application to standard image recognition datasets, recent works demonstrate the vulnerability of widely-used gradient-based interpretation schemes to norm-bounded perturbations adversarially designed for every individual input sample. However, such adversarial perturbations are commonly designed using the knowledge of an input sample, and hence perform sub-optimally in application to an unknown or constantly changing data point. In this paper, we show the existence of a \emph{Universal Perturbation for Interpretation (UPI)} for standard image datasets, which can alter a gradient-based feature map of neural networks over a significant fraction of test samples. To design such a UPI, we propose a gradient-based optimization method as well as a principal component analysis (PCA)-based approach to compute a UPI which can effectively alter a neural network's gradient-based interpretation on different samples. We support the proposed UPI approaches by presenting several numerical results of their successful applications to standard image datasets.

%% file: 1-introduction.tex
Deep neural networks (DNNs) have attained impressive results in many computer vision \cite{krizhevsky2017imagenet}, speech processing \cite{deng2013new}, and computational biology \cite{alipanahi2015predicting} problems. While DNNs usually achieve state-of-the-art performance in application to image and speech datasets, their performance is highly vulnerable to minor perturbations adversarially designed for an input sample \cite{szegedy2013intriguing,goodfellow2014explaining}. Recently, it has been shown that not only do adversarial perturbations alter the prediction of DNN machines, but further one can design small norm-bounded perturbations to significantly change the gradient-based feature maps used for interpreting the prediction of neural networks \cite{2-3,heo2019fooling}. Understanding the mechanisms behind such adversarial attacks has received significant attention in the recent machine learning literature.

Similar to standard adversarial attacks, the adversarial perturbations targeting the gradient-based interpretation of DNNs are commonly designed separately for every input sample, which requires full knowledge of the input data point. However, the input sample can constantly alter in several adversarial attack scenarios, and it may not be computationally feasible to optimize different perturbations for a continuously-changing input data point. Furthermore, in several real-world adversarial attack scenarios, the input features are hidden in test time, and therefore the adversary has no or little information of the target input sample. In these adversarial attack problems, the adversary can still design and apply a universal adversarial perturbation (UAP) \cite{moosavi2017universal} to alter the gradient-based interpretation for a typical input sample. While several UAP attack schemes \cite{moosavi2017universal,poursaeed2018generative,hayes2018learning,khrulkov2018art,deshpande2019universal,behjati2019universal,xie2020real,zhang2021attack,liu2022efficient} have been proposed to target the prediction of neural networks in the context of classification, the current deep learning literature lacks universal attack schemes that target the gradient-based interpretation of neural networks.

In this paper, we focus on designing \emph{Universal Perturbations for Interpretation (UPI)} as universal attacks aimed to change the saliency maps of neural nets over a significant fraction of input data. To achieve this goal, we formulate an optimization problem to find a UPI perturbation with the maximum impact on the total change in the gradient-based feature maps over the training samples. We propose a projected gradient method called UPI-Grad for solving the formulated optimization problem.
Furthermore, in order to handle the difficult non-convex nature of the formulated optimization problem, we develop a principal component analysis (PCA)-based approach called UPI-PCA to approximate the solution to this problem using the top singular vector of fast gradient method (FGM) perturbations to the interpretation vectors. We demonstrate that the spectral UPI-PCA scheme yields the first-order approximation of the solution to the UPI-Grad optimization problem.

To implement the UPI-PCA scheme for generating universal perturbations, we propose a stochastic optimization method which can efficiently converge to the top singular vector of first-order interpretation-targeting perturbations. 
Finally, we demonstrate our numerical results of applying the UPI-Grad and UPI-PCA methods to standard image recognition datasets and neural network architectures. Our numerical results reveal the vulnerability of commonly-used gradient-based feature maps to universal perturbations which can significantly alter the interpretation of neural networks. The empirical results show the satisfactory convergence of the proposed stochastic optimization method to the top singular vector of the attack scheme, and further indicate the proper generalization of the designed attack vector to test samples unseen during the optimization of the universal perturbation. We can summarize the contributions of this work as follows:
\begin{itemize}[leftmargin=*]
    \item Extending universal adversarial attacks to alter gradient-based interpretations of neural networks,
    \item Proposing the gradient-based UPI-Grad method to design universal perturbations for the neural nets' interpretation,
    \item Developing the spectral UPI-PCA method for optimizing universal perturbations to the interpretation of neural networks,
    \item Providing numerical evidence on the sensitivity of the interpretation of neural networks to universal perturbations. 
\end{itemize}

%% file: 2-prelim.tex
\subsection{Interpretation methods}

In this section, we review standard gradient-based interpretation mechanisms for deep neural network classifiers and explain the notation and definitions used in the paper. 
Throughout the paper, we use notation $X\in\mathcal{X}\subseteq\mathbb{R}^d$ to denote the random vector of input features and notation $Y\in\mathcal{Y}=\{1,2,\ldots,k\}$ to denote a $k$-ary random label which the neural network classifier aims to predict. Here, the goal of the deep neural network learner is to find a function $f\in\mathcal{F}_{\text{\rm nn}}$ from neural net space $\mathcal{F}_{\text{\rm nn}}$ which can accurately predict $Y$ from an observation of $X$. To do this, we apply the standard empirical risk minimization (ERM) approach which minimizes the empirical expected loss $\mathbb{E}[\ell(f(X),Y)]$ over training samples $(x_i,y_i)_{i=1}^n$ and loss function $\ell(y,\hat{y})$ between predicted label $\hat{y}$ and actual label $y$.

For the interpretation of a trained neural net $f(x)$ at an input $x\in\mathbb{R}^d$, we use feature importance maps as an explanation of the classifier $f$'s prediction. Given the neural net’s predication $c\in\mathcal{Y}$ for the input $x$ (the label with the maximum prediction score), we use $S_c(x)$, to denote the output of the $f$ final layer's neuron corresponding to label $c$. 
Here, we briefly review two widely-used gradient-based interpretation schemes for generating a feature importance map, which we later use in our numerical analysis.

\begin{itemize}[leftmargin=*]
    \item \textbf{Simple gradient method:}
    As introduced in \cite{2-1} and further analyzed in \cite{2-2} for deep neural network classifiers, the simple gradient method is based on a first-order linear approximation of the classifier neural network's output and is defined as the following normalized gradient vector:
    \begin{equation}
        I_{\textrm{SG}} (x) \, := \, \frac{\nabla_x S_c(x)}{\bigl\Vert \nabla_x S_c(x)\bigr\Vert_1}
        \label{l3}
    \end{equation}
    In the above, $\Vert \cdot\Vert_1$ denotes the $\ell_1$-norm, i.e. the summation of the absolute value of an input vector's entries.  
    \item \textbf{Integrated gradients method:}
    As introduced in \cite{2-3}, the feature importance map of the integrated gradient method is defined as the $\ell_1$-normalized version of the following score:
    \begin{align}
        I^{\textrm{\tiny unnorm}}_{\textrm{IG}} (x) \, :=& \,  \frac{\Delta x}{M} \sum_{k=1}^{M} \nabla_x S_c\bigl(x^{0}+ \frac{k}{M}\Delta x \bigr) , \nonumber \\
        I_{\textrm{IG}} (x)  \, =&\, \frac{I^{\textrm{\tiny unnorm}}_{\textrm{IG}}(x) }{\bigl\Vert I^{\textrm{\tiny unnorm}}_{\textrm{IG}}(x)  \bigr\Vert_1}      
        \label{l4}
    \end{align}
    In this definition, $M$ is the number of intermediate points between a reference point $x^0$ and input $x$, and $  \Delta x = x - x^{0}$ is the difference between the input and reference points.
\end{itemize}

\subsection{Adversarial Attacks and Universal Perturbations}
Given a neural network function $f$, an $\epsilon$-norm-bounded adversarial perturbation is a vector $\delta$ satisfying $\Vert \delta \Vert \le \epsilon$ which maximizes the classification loss function in an $\epsilon$ distance from an input $(x,y)$, i.e. the solution to the following optimization problem:
\begin{equation}
    \max_{\delta:\: \Vert \delta\Vert \le \epsilon}\; \ell\bigl(y,f(x+\delta) \bigr).
\end{equation}
\cite{2-3} extends the concept of an adversarial perturbation to neural networks' interpretations. Here, for an interpretation scheme $I(x)$, the aim is to find a norm-bounded vector $\delta\in\mathbb{R}^d$ that maximizes a distance function $\mathcal{D}(\cdot,\cdot)$ between the original and perturbed data points, 
\begin{equation}
    \max_{\delta\in\mathbb{R}^d:\: \Vert \delta\Vert \le \epsilon}\; \mathcal{D}\bigl( I(x),I(x+\delta) \bigr).
\end{equation}

While standard adversarial attacks assign different perturbation vectors to different input data, universal adversarial perturbations (UAPs) introduced in \cite{moosavi2017universal} use the same perturbation vector for every input sample. Therefore, UAPs provide a more effective approach to alter neural net predictions for an unknown input. A standard approach for designing classification-based UAPs is to solve the following optimization problem for a set of training data $\{(x_i,y_i)_{i=1}^n\}$:
\begin{equation}
     \max_{\delta:\: \Vert \delta\Vert \le \epsilon}\; \frac{1}{n}\sum_{i=1}^n\ell\bigl(y_i,f(x_i+\delta) \bigr).
\end{equation}

%% file: 3-algorithm.tex
As discussed in the previous sections, while universal perturbations are extensively studied in the context of classification, the current literature lacks a universal perturbation scheme that targets the neural net's gradient-based interpretation. Such a universal perturbation for the interpretation of neural nets will be useful to target the gradient-based interpretation of a neural network for an unobserved or constantly-updating input sample.

\begin{figure*}[!h]
    \centering
    \small
    \stackunder[0pt]{
    \includegraphics[width=0.246\columnwidth]{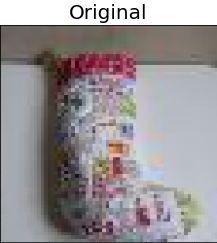}
    \includegraphics[width=0.246\columnwidth]{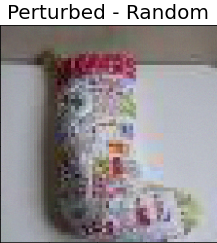}
    \includegraphics[width=0.246\columnwidth]{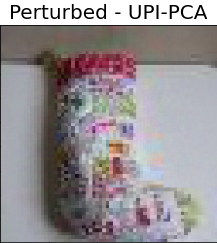}
    \includegraphics[width=0.246\columnwidth]{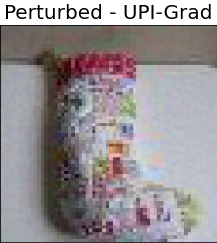}
    }{}
    \stackunder[0pt]{
    \includegraphics[width=0.246\columnwidth]{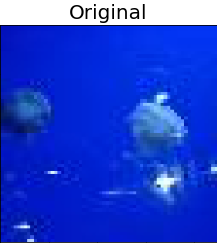}
    \includegraphics[width=0.246\columnwidth]{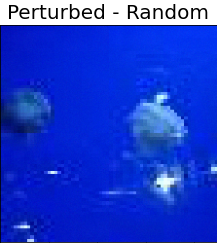}
    \includegraphics[width=0.246\columnwidth]{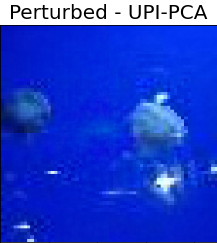}
    \includegraphics[width=0.246\columnwidth]{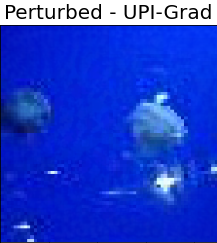}
    }{}
    \\
    \vspace{2pt}
    \stackunder[5pt]{
    \includegraphics[width=0.246\columnwidth]{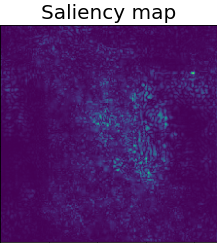}
    \includegraphics[width=0.246\columnwidth]{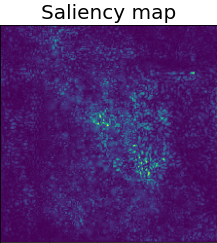}
    \includegraphics[width=0.246\columnwidth]{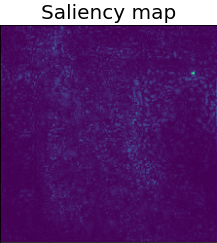}
    \includegraphics[width=0.246\columnwidth]{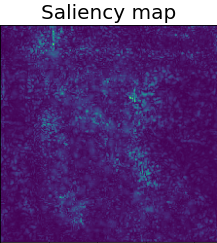}
    }{(a) Simple Gradient}
    \stackunder[5pt]{
    \includegraphics[width=0.246\columnwidth]{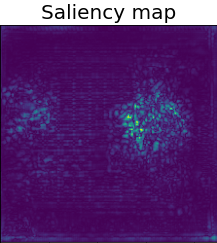}
    \includegraphics[width=0.246\columnwidth]{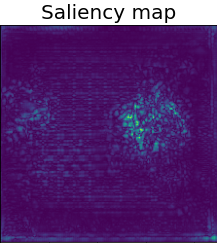}
    \includegraphics[width=0.246\columnwidth]{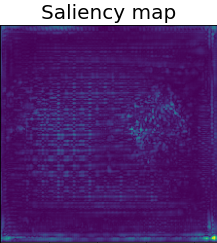}
    \includegraphics[width=0.246\columnwidth]{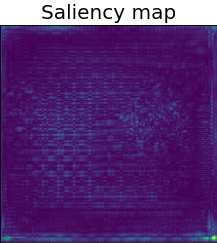}
    }{(b) Integrated Gradients}
    \caption{Interpretation of  VGG-16 on two Tiny-ImageNet samples before and after adding the perturbations.}
    \label{Figure:Visulaization_Interpretation}\vspace{-3mm}
\end{figure*}

Here, we propose \emph{Universal Perturbations for Interpretation (UPI)} which is a universal attack scheme targeting gradient-based interpretations of deep neural net classifiers. To define the optimization problem for UPI, we propose the following maximization problem for interpretation scheme $I$ and distance measure $\mathcal{D}$:
\begin{align}\label{UPI Optimization: Original}
\begin{aligned}
& \underset{\delta\in\mathbb{R}^d}{\arg\!\max}\quad \mathbb{E}_{Z\sim \mathcal{N}(\mathbf{0},\sigma^{2} I)}\biggl[ \frac{1}{n}\sum_{i = 1}^{n} \mathcal{D} \bigl(I(x_i), I(x_i + \delta + Z)\bigr)\biggr] \\ & \textrm{subject to } \quad \|\delta\| \le \epsilon
\end{aligned}
\end{align}
Here, we search for a norm-bounded perturbation $\delta\in\mathbb{R}^d$ that results in the maximum change in the summation of distances between unperturbed and universally-perturbed interpretations. Note that we define the objective function as the expectation over an additive Gaussian $Z\sim \mathcal{N}(\mathbf{0},\sigma^{2} I)$ with a zero mean and an isotropic covariance matrix $\sigma^2 I$ to smoothen the objective function and address the zero gradient of the  distance function at the zero point $\delta =\mathbf{0}$. Note that $\sigma\ge 0 $ is a hyperparameter in the UPI framework which we choose to be significantly smaller than $\sigma\ll \epsilon$ in our numerical experiments. To solve the above optimization problem, we propose a projected gradient ascent method described in Algorithm \ref{alg:upi-grad}.

However, we note that similar to the classification-based universal perturbations, the above optimization problem is in general non-convex and can be significantly challenging as a local search method could get stuck at sub-optimal stationary points due to the existence of multiple locally optimal perturbations. As discussed in \cite{khrulkov2018art,deshpande2019universal}, a significant challenge in such optimization problems is the symmetric behavior of the objective function around the zero perturbation $\delta=\mathbf{0}$, which can lead to the failure of a gradient-based method in solving the optimization task.

  To address this challenge, we propose a variant of the optimization problem in \eqref{UPI Optimization: Original}, which only concerns finding the optimal perturbation direction. Here, our goal is to find a universal direction $\delta\in\mathbb{R}^d$ which is aligned with the universally-aligned perturbations to input samples. In the following, we design such an optimization problem where the perturbations are all aligned with the universal direction $\delta$ while they could have different magnitudes chosen by the sample-based scalar variable $c_i$'s. We also regularize $c_i$'s to have a bounded absolute value through the regularization term $-\frac{\lambda}{2}c_i^2$.\vspace{-2mm}
\begin{align}\label{UPI Optimization: Universal Direction}
& \underset{\delta\in\mathbb{R}^d}{\max}\quad \frac{1}{n}\sum_{i = 1}^{n} \max_{c_i \in\mathbb{R}}\: \mathbb{E}\biggl[\mathcal{D} \bigl(I(x_i), I(x_i + Z + c_i  \delta)\bigr) -\frac{\lambda}{2}c_i^2 \biggr] \nonumber \\ & \textrm{subject to } \quad \|\delta\| \le  \epsilon
\end{align}

\begin{algorithm}[t]
\caption{UPI-Grad}
\label{alg:upi-grad}
\begin{algorithmic} 
\STATE \textbf{Input:} Training data $(x_i,y_i)_{i=1}^n$, maximum perturbation norm $\epsilon$, feature map $I(\cdot)$, stepsize $\alpha$, standard deviation $\sigma$
\STATE \textbf{Define} dissimilarity function $\mathcal{D} (x, x') = \|I(x) - I(x')\|^{2}_{2}$
\STATE \textbf{Initialize} $\delta=z$ with a normal $z\sim\mathcal{N}(\mathbf{0},\sigma^2 I)$

\FOR{epoch $\in \ \{1, \dots, N_{ep}\}$}
\FOR{minibatch $B \subset X_{1:n} $}
\STATE \textbf{Draw} normal $z\sim\mathcal{N}(\mathbf{0},\sigma^2 I)$
\STATE \textbf{Update} $\delta \leftarrow \delta + \frac{\alpha}{|B|} \sum_{x\in B} \nabla_\delta \mathcal{D}(x, x +z + \delta )$ 
\STATE \textbf{Project} on the $\epsilon$-norm ball $\delta = \frac{\delta}{\max\{1,{\Vert \delta\Vert}/{\epsilon} \} }$ 
\ENDFOR
\ENDFOR
\STATE \textbf{Output:} Universal perturbation $\delta $
\end{algorithmic}
\end{algorithm}

To have a first-order approximation of the solution to the above problem, we use the following proposition:
\begin{prop}\label{Prop: 1}
Consider the objective function in \eqref{UPI Optimization: Universal Direction}. Suppose that every summation term $\mathcal{D}(I(x),I(x+\delta))$ is $L$-Lipschitz in $\delta$. 
Then, assuming that $\tau:=\frac{L\sqrt{d}\epsilon^2}{\sigma} <\lambda $, for every $\Vert\delta\Vert_2\le\epsilon$ we will have
\begin{align*}
&\frac{1}{\lambda+\tau} \left(\delta^\top \mathbb{E}\bigl[\nabla_z\mathcal{D} \bigl(I(x_i), I(x_i + Z)\bigr) \bigr]\right)^2 \\
\le\, &\biggl\{\max_{c_i \in\mathbb{R}}\: \mathbb{E}\left[\mathcal{D} \bigl(I(x_i), I(x_i + Z + c_i  \delta)\bigr)\right] -\frac{\lambda}{2}c_i^2 \biggr\} \\
&  - \mathbb{E}\left[\mathcal{D} \bigl(I(x_i), I(x_i + Z )\bigr)\right]  \\
\le \, &\frac{1}{\lambda-\tau}
\left(\delta^\top \mathbb{E}\bigl[\nabla_z\mathcal{D} \bigl(I(x_i), I(x_i + Z)\bigr) \bigr]\right)^2.
\end{align*}
\end{prop}
\begin{proof}
We defer the proof to the Appendix of \cite{completepaper}.
\end{proof}
Based on the above proposition, we obtain the following optimization problem as the first-order approximation of \eqref{UPI Optimization: Universal Direction}:
 \begin{align}\label{UPI_FO_PCA}
\begin{aligned}
& \underset{\delta\in\mathbb{R}^d}{\max}\quad \frac{1}{n}\sum_{i = 1}^{n}\, \left(\delta^\top \mathbb{E}\bigl[\nabla_z\mathcal{D} \bigl(I(x_i), I(x_i + Z)\bigr) \bigr]\right)^2  \\ & \textrm{subject to } \quad \|\delta\| \le  \epsilon
\end{aligned}
\end{align}
The next proposition reveals that the solution to the above problem is indeed the top singular vector of the first-order gradients of the randomly-perturbed distance function at $\delta=\mathbf{0}$:
\begin{prop}\label{Prop: 2}
Consider the optimization problem in Equation \eqref{UPI_FO_PCA} for the $\ell_2$-norm case. Then, the solution to the optimization problem is the top right singular vector of the following matrix:
\begin{equation*}
    G := \mathbb{E}_{Z\sim\mathcal{N}(\mathbf{0},\sigma^2 I)}\biggl[\, \begin{bmatrix} \nabla_z\mathcal{D} \bigl(I(x_1), I(x_1 + Z)\bigr) \\
    \vdots \\
    \nabla_z\mathcal{D} \bigl(I(x_n), I(x_n + Z)\bigr) 
    \end{bmatrix} \, \biggr].
\end{equation*}
\end{prop}
\begin{proof}
We defer the proof to the Appendix of \cite{completepaper}.
\end{proof}
The above result suggests using the top principal component of the gradient matrix $G$ as the UPI perturbation. Hence, we propose the principal component analysis (PCA)-based UPI-PCA in Algorithm~\ref{alg:upi-pca} as a stochastic power method for computing the top right singular vector of  matrix $G$. 

\begin{algorithm}[t]
\caption{UPI-PCA}
\label{alg:upi-pca}
\begin{algorithmic} 
\STATE \textbf{Input:} Training data $(x_i,y_i)_{i=1}^n$, maximum perturbation norm $\epsilon$, feature map $I(\cdot)$, stepsize $\alpha$, standard deviation $\sigma$
\STATE \textbf{Define} dissimilarity function $\mathcal{D} (x, x') = \|I(x) - I(x')\|^{2}_{2}$
\STATE \textbf{Initialize} $\delta=z$ with a normal $z\sim\mathcal{N}(\mathbf{0},\sigma^2 I)$
\FOR{epoch $\in \ \{1, \dots, N_{ep}\}$}
\FOR{minibatch $B \subset X_{1:n} $}
\STATE \textbf{Draw} random $z\sim\mathcal{N}(\mathbf{0},\sigma^2 I)$
\STATE \textbf{Compute} gradients $G = \left[{\nabla_{z}\mathcal{D}(x, x +z)}\right]_{x\in B} $
\STATE \textbf{Update} $\delta \leftarrow  \delta + \frac{\alpha}{|B|}\sum_{i=1}^{|B|}\bigl(\delta^\top G_i\bigr)G_i $ 
\STATE \textbf{Project} on the $\epsilon$-norm ball $\delta = \frac{\delta}{\max\{1,{\Vert \delta\Vert}/{\epsilon} \} }$ 
\ENDFOR
\ENDFOR
\STATE \textbf{Output:} Universal perturbation $\delta$
\end{algorithmic}
\end{algorithm}

\begin{table*}
\begin{center}
\caption{Average dissimilarity of VGG-16, MobileNet, and 2-layer CNN trained on Tiny-ImageNet, CIFAR-10, and MNIST respectively.}
\label{table:table1}
\resizebox{0.95\textwidth}{!}{%
\begin{tabular}{lllcc|cc|ccc}
\toprule
{} & {} & {} & \multicolumn{2}{c}{\textbf{Tiny-ImageNet}} & \multicolumn{2}{c}{\textbf{CIFAR-10}} & \multicolumn{3}{c}{\textbf{MNIST}} \\\\
Interpretation method & Attack & Type & VGG-16 & MobileNet & VGG-16 & MobileNet & MobileNet & 2-layer CNN & 2-layer FCN\\
\midrule
\multirow{8}{*}{Simple Gradient} & \multirow{2}{*}{Random} & Per image & 0.407 & 0.664 & 0.275 & 0.482 & 0.342 & 1.012 & 0.085 \\
\cmidrule{3-10}
{} & {} & Universal & 0.407 & 0.665 & 0.274 & 0.485 & 0.342 & 1.006 & 0.079 \\
\cmidrule{2-10}
{} & \multirow{5}{*}{Interpretation} & Per image-PGD & 0.777 & 0.783 & 0.590 & 0.710 & 0.566 & 1.024 & 0.324 \\
{} & {} & Per image-FGM & 0.772 & 0.778 & 0.573 & 0.705 & 0.549 & 0.911 & 0.334 \\
\cmidrule{3-10}
{} & {} & UPI-PCA-PGD (Ours) & 0.550 & 0.681 & 0.350 & 0.540 & 0.428 & \textbf{1.501} & 0.308 \\
{} & {} & UPI-PCA-FGM (Ours) & 0.538 & 0.668 & 0.343 & 0.553 & 0.428 & 1.275 & \textbf{0.358} \\
{} & {} & UPI-Grad (Ours) & \textbf{0.598} & \textbf{0.719} & \textbf{0.439} & \textbf{0.641} & \textbf{0.441} & 1.045 & 0.242 \\
\cmidrule{2-10}
{} & Classification & Universal & 0.540 & 0.699 & 0.372 & 0.587 & 0.379 & 1.042 & 0.185 \\
\midrule
\multirow{8}{*}{Integrated Gradients} & \multirow{2}{*}{Random} & Per image & 0.278 & 0.528 & 0.201 & 0.312 & 0.240 & 0.798 & 0.028 \\
\cmidrule{3-10}
{} & {} & Universal & 0.277 & 0.528 & 0.199 & 0.311 & 0.241 & 0.794 & 0.026 \\
\cmidrule{2-10}
{} & \multirow{5}{*}{Interpretation} & Per image-PGD & 0.749 & 0.754 & 0.560 & 0.652 & 0.510 & 0.828 & 0.246 \\
{} & {} & Per image-FGM & 0.734 & 0.728 & 0.531 & 0.618 & 0.476 & 0.768 & 0.228 \\
\cmidrule{3-10}
{} & {} & UPI-PCA-PGD (Ours) & 0.430 & 0.578 & 0.271 & 0.431 & 0.318 & 0.849 & 0.166 \\
{} & {} & UPI-PCA-FGM (Ours) & 0.434 & 0.561 & 0.258 & 0.490 & 0.321 & \textbf{0.851} & 0.166 \\ 
{} & {} & UPI-Grad (Ours) & \textbf{0.531} & \textbf{0.676} & \textbf{0.287} & \textbf{0.541} & \textbf{0.385} & 0.835 & \textbf{0.183} \\
\cmidrule{2-10}
{} & Classification & Universal & 0.408 & 0.581 & 0.280 & 0.412 & 0.272 & 0.824 & 0.063 \\
\bottomrule
\end{tabular}
}
\end{center}\vspace{-3mm}
\end{table*}

%% file: 4-numericals.tex
\begin{figure*}[h!]
    \centering
    \small
    \stackunder[0pt]{
    \stackunder[0pt]{\includegraphics[width=0.135\textwidth]{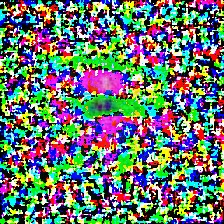}}{} \hspace{5mm}
    \>
    \stackunder[0pt]{\includegraphics[width=0.135\textwidth]{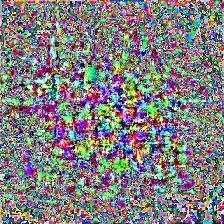}}{}
    }{}
    \hspace{2cm}
    \stackunder[0pt]{
    \stackunder[0pt]{\includegraphics[width=0.135\textwidth]{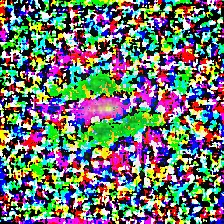}}{}\hspace{5mm}
    \>
    \stackunder[0pt]{\includegraphics[width=0.135\textwidth]{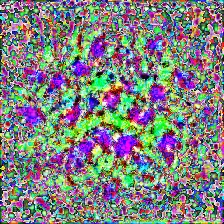}}{}
    }{}
    \\
    \vspace{3pt}
    \stackunder[5pt]{
    \stackunder[3pt]{\includegraphics[width=0.135\textwidth]{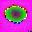}}{UPI-PCA}\hspace{5mm}
    \>
    \stackunder[3pt]{\includegraphics[width=0.135\textwidth]{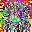}}{UPI-Grad}
    }{Simple Gradient}
    \hspace{2cm}
    \stackunder[5pt]{
    \stackunder[3pt]{\includegraphics[width=0.135\textwidth]{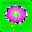}}{UPI-PCA}\hspace{5mm}
    \>
    \stackunder[3pt]{\includegraphics[width=0.135\textwidth]{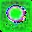}}{UPI-Grad}
    }{Integrated Gradients}
    \caption{Visualization of UPI perturbations in VGG-16 experiments. The top and bottom rows are the Tiny-ImageNet and CIFAR-10 UPIs.}\vspace{-4mm}
    \label{Figure:Visulaization_Perturbation}
\end{figure*}

We performed numerical experiments to evaluate the proposed UPI schemes on standard image datasets including MNIST \cite{lecun1998mnist}, CIFAR-10 \cite{krizhevsky2009learning}, and Tiny-ImageNet \cite{imagenet_cvpr09}. Note that Tiny-ImageNet is a downsized version of the ImageNet dataset containing 100,000 images from 200 ImageNet classes, with 500 images per class. For universal attacks against feature importance interpretation, we trained VGG-16 \cite{simonyan2014very} and MobileNet \cite{howard2017mobilenets} models on Tiny-ImageNet and CIFAR-10 datasets. For MNIST, we trained a MobileNet, a simple 2-layer convolution neural net (CNN), and a simple 2-layer fully-connected net (FCN) with the ReLU activation.

In the experiments, we used the following normalzied dissimilarity score to evaluate the performance of the perturbations:
\begin{equation}
    \mathcal{D} (x, x') := \frac{\|I(x) - I(x')\|_{2}}{\|I(x)\|_{2}}
\end{equation}
where $x,x'$ denote the original and perturbed images, and $I(\cdot)$ is the specified interpretation scheme. 
We report the mean of the above dissimilarity scores over the test set. All the perturbations were scaled such that the norm value $\epsilon = \frac{1000}{256}$ for Tiny-ImageNet, and $\epsilon = \frac{100}{256}$ for CIFAR-10 and MNIST. Since the resolution of CIFAR-10 and MNIST samples were lower than Tiny-ImageNet, the norm value was chosen to be smaller. 
We ran all per-image iterative attack algorithms for $150$ iterations, and stepsize $\alpha = 0.5$.  
We evaluated the random attacks 10 times with different seeds, and report the results with the mean dissimilarity value.

As reported in Table \ref{table:table1}, we compared UPIs with several baselines including (i) a normally-distributed random   perturbation generated both per-image and universally, (ii) per-image adversarial perturbations on interpretation designed individually for each input image using the PGD method in \cite{2-3}, (iii) classification-based universal adversarial perturbation as described in \cite{4-2}. We also report the dissimilarity scores for the proposed UPI-Grad  and UPI-PCA algorithms.

As Table \ref{table:table1}'s scores suggest, the UPIs performed significantly better than a random perturbation, but were often a little weaker than per-image adversarial perturbations. In some cases, UPIs even performed as well as per-image adversarial perturbations. 
In addition, the average dissimilarity of classification-based  UAPs is shown in Table \ref{table:table1}. While their achieved scores were somewhat inferior to UPI-Grad,  they still performed successfully in comparison to random and per-image perturbations.  Figure~\ref{Figure:Visulaization_Interpretation} visually shows the effect of the designed perturbations on the interpretations of VGG-16 classifiers. Also, Figure~\ref{Figure:Visulaization_Perturbation} visualizes the designed UPIs for the trained VGG-16-nets and highlights their visual patterns. These empirical results confirm our hypothesis that universal perturbations (both UAPs and UPIs) can significantly change the interpretation of neural networks.


%% file: 5-appendix.tex
\section{Proofs}
\subsection{Proof of Proposition \ref{Prop: 1}}
To prove this proposition, we first review Stein's lemma \cite{landsman2008stein} from the statistics literature.
\begin{lemma}[Stein's lemma \cite{landsman2008stein}]
 For every Lipschitz-continuous function $f:\mathbb{R}^d\rightarrow\mathbb{R}$ and isotropic Gaussian vector $Z\sim\mathcal{N}(\mathbf{0},\sigma^2 I)$, the following equality holds:
\begin{equation*}
    \mathbb{E}\bigl[ \nabla f(x+Z) \bigr] = \mathbb{E}\bigl[ \frac{f(x+Z)}{\sigma^2} {Z}\bigr].
\end{equation*}
\end{lemma}
\noindent Using Stein's lemma, for every $x,x',y\in\mathbb{R}^d$ we have
\begin{align*}
    & \bigl\Vert \nabla_x \mathbb{E}[\mathcal{D}(y,x+Z)] - \nabla_x \mathbb{E}[\mathcal{D}(y,x'+Z)] \bigr\Vert_2 \\
     =\, & \bigl\Vert  \mathbb{E}[\nabla_x\mathcal{D}(y,x+Z)] -  \mathbb{E}[\nabla_x\mathcal{D}(y,x'+Z)] \bigr\Vert_2 \\
    \stackrel{(a)}{=}\, & \left\Vert \mathbb{E}\left[ \frac{Z}{\sigma^2} \mathcal{D}(y,x+Z)\right] -\mathbb{E}\left[\frac{Z}{\sigma^2} \mathcal{D}(y,x'+Z) \right] \right\Vert_2 \\
    =\, & \left\Vert \mathbb{E}\biggl[ \frac{Z}{\sigma^2} \bigl( \mathcal{D}(y,x+Z) - \mathcal{D}(y,x'+Z)\bigr)\biggr] \right\Vert_2 \\
    \stackrel{(b)}{\le}\, &  \mathbb{E}\biggl[ \left\Vert\frac{Z}{\sigma^2} \bigl( \mathcal{D}(y,x+Z) - \mathcal{D}(y,x'+Z)\bigr)\right\Vert_2\biggr]  \\
    =\, &  \mathbb{E}\biggl[ \frac{\Vert Z\Vert_2}{\sigma^2} \bigl\vert \mathcal{D}(y,x+Z) - \mathcal{D}(y,x'+Z)\bigr\vert\biggr]  \\
    \stackrel{(c)}{\le}\, &  \mathbb{E}\biggl[ \frac{\Vert Z\Vert_2}{\sigma^2} L \Vert x-x'\Vert_2 \biggr] \\
    =\, & \frac{L\Vert x-x'\Vert_2}{\sigma^2} \mathbb{E}\bigl[ {\Vert Z\Vert_2}  \bigr] \\
    \stackrel{(d)}{=} \, & \frac{L\Vert x-x'\Vert_2}{\sigma^2} \sqrt{\frac{2d\sigma^2}{\pi}} \\
    \le \, & \frac{L\sqrt{d}\Vert x-x'\Vert_2}{\sigma}.
\end{align*}
In the above, (a) follows from Stein's lemma. (b) comes from the application of Jensen's inequality to the convex norm  function $\Vert\cdot\Vert_2$. (c) is the consequence of the proposition's assumption on the Lipschitz coefficient of $\mathcal{D}$. $(d)$ uses the analytical solution for the expected value $\mathbb{E}\bigl[ {\Vert Z\Vert_2}  \bigr]$ according to $Z\sim \mathcal{N}(\mathbf{0},\sigma^2 I_{d\times d})$, which is $\sqrt{{2d\sigma^2}/{\pi}}$.

Therefore, the gradient of $\mathbb{E}[\mathcal{D}(y,x+Z)]$ with respect to $x$ will be $\frac{L\sqrt{d}}{\sigma}$-Lipschitz, which means that $\mathbb{E}[ \mathcal{D}(y,x+Z)]$ is $\frac{L\sqrt{d}}{\sigma}$-smooth in $x$ and satisfies the following inequality for every $x,x',y$:
\begin{align*}
 & \nabla_x \mathbb{E}[\mathcal{D}(y,x+Z)]^\top ({x}'-{x})-\frac{L\sqrt{d}}{2\sigma} \Vert x -x' \Vert^2_2\\
 \le \;\; &\mathbb{E}[\mathcal{D}(y,x'+Z)] - \mathbb{E}[\mathcal{D}(y,x+Z)] \\
    \le\;\; & \nabla_x \mathbb{E}[\mathcal{D}(y,x+Z)]^\top ({x}'-{x}) + \frac{L\sqrt{d}}{2\sigma} \Vert x -x' \Vert^2_2.
\end{align*}
As  a result, for any $c_i\in\mathbb{R}$ and $x_i,\delta\in\mathbb{R}^d$ the following holds:
\begin{align*}
    & c_i \nabla_x \mathbb{E}[\mathcal{D}(x_i,x_i+Z)]^\top \delta - \frac{L\sqrt{d} c_i^2}{2\sigma} \Vert \delta \Vert^2_2\\
    \le\;\; & \mathbb{E}[\mathcal{D}(x_i,x_i+Z +c_i\delta )] - \mathbb{E}[\mathcal{D}(x_i,x_i+Z  )]  \\
    \le\;\;   & c_i \nabla_x \mathbb{E}[\mathcal{D}(x_i,x_i+Z)]^\top \delta + \frac{L\sqrt{d} c_i^2}{2\sigma} \Vert \delta \Vert^2_2.
\end{align*}
Thus, assuming that $\Vert \delta\Vert_2\le \epsilon$ and defining $\tau = \frac{L\sqrt{d}\epsilon^2}{\sigma}$, we will have the following inequalities
\begin{align*}
    & c_i \nabla_x \mathbb{E}[\mathcal{D}(x_i,x_i+Z)]^\top \delta - \frac{\tau  }{2} c_i^2 \\
    \le\;\; & \mathbb{E}[\mathcal{D}(x_i,x_i+Z +c_i\delta )] -  \mathbb{E}[\mathcal{D}(x_i,x_i+Z  )]  \\
    \le\;\;   & c_i \nabla_x \mathbb{E}[\mathcal{D}(x_i,x_i+Z)]^\top \delta + \frac{\tau  }{2}c_i^2,
\end{align*}
which implies that
\begin{align*}
    &  c_i \nabla_x \mathbb{E}[\mathcal{D}(x_i,x_i+Z)]^\top \delta - \frac{\lambda + \tau  }{2} c_i^2\\
    \le\;\; & \mathbb{E}[\mathcal{D}(x_i,x_i+Z +c_i\delta )] -\frac{\lambda}{2}c_i^2 - \mathbb{E}[\mathcal{D}(x_i,x_i+Z  )] \\
    \le\;\;   & c_i \nabla_x \mathbb{E}[\mathcal{D}(x_i,x_i+Z)]^\top \delta - \frac{\lambda - \tau  }{2}c_i^2.
\end{align*}
The assumption $\lambda>\tau$ implies that the above upper and lower bounds are concave quadratic functions of $c_i$. Therefore, maximizing the sides of the above inequality over $c_i\in\mathbb{R}$ shows that 
\begin{align*}
&\frac{1}{\lambda+\tau} \left(\delta^\top \mathbb{E}\bigl[\nabla_z\mathcal{D} \bigl(I(x_i), I(x_i + Z)\bigr) \bigr]\right)^2 \\
\le\, &\biggl\{\max_{c_i \in\mathbb{R}}\: \mathbb{E}\left[\mathcal{D} \bigl(I(x_i), I(x_i + Z + c_i  \delta)\bigr)\right] -\frac{\lambda}{2}c_i^2 \biggr\} \\
&  - \mathbb{E}\left[\mathcal{D} \bigl(I(x_i), I(x_i + Z )\bigr)\right]  \\
\le \, &\frac{1}{\lambda-\tau}
\left(\delta^\top \mathbb{E}\bigl[\nabla_z\mathcal{D} \bigl(I(x_i), I(x_i + Z)\bigr) \bigr]\right)^2,
\end{align*}
which completes the proof.
\subsection{Proof of Proposition \ref{Prop: 2}}
Note that the objective function in the target optimization problem can be written as:
\begin{align*}
   &\frac{1}{n}\sum_{i = 1}^{n}\, \left(\delta^\top \mathbb{E}\bigl[\nabla_z\mathcal{D} \bigl(I(x_i), I(x_i + Z)\bigr) \bigr]\right)^2 \\
  =\, & \frac{1}{n}\sum_{i = 1}^{n}\biggl[\, \left(\delta^\top \mathbb{E}\bigl[\nabla_z\mathcal{D} \bigl(I(x_i), I(x_i + Z)\bigr) \bigr]\right)\\
  &\qquad \times \left( \mathbb{E}\bigl[\nabla_z\mathcal{D} \bigl(I(x_i), I(x_i + Z)\bigr) \bigr]^\top \delta\right)\biggr]\\
  =\, & \frac{1}{n}\sum_{i = 1}^{n}\biggl[\, \delta^\top \mathbb{E}\bigl[\nabla_z\mathcal{D} \bigl(I(x_i), I(x_i + Z)\bigr) \bigr] \\
  &\qquad \times \mathbb{E}\bigl[\nabla_z\mathcal{D} \bigl(I(x_i), I(x_i + Z)\bigr) \bigr]^\top \delta\biggr]\\ 
  =\, & \delta^\top\biggl( \frac{1}{n}\sum_{i = 1}^{n}\biggl[ \mathbb{E}\bigl[\nabla_z\mathcal{D} \bigl(I(x_i), I(x_i + Z)\bigr) \bigr]\\
  &\qquad \times \mathbb{E}\bigl[\nabla_z\mathcal{D} \bigl(I(x_i), I(x_i + Z)\bigr) \bigr]^\top \biggr] \biggr) \delta\\ 
  =\, &\delta^\top M \delta
\end{align*}
where we define matrix $M$ as
\begin{equation*}
    M := \frac{1}{n}\sum_{i = 1}^{n}\biggl[ \mathbb{E}\bigl[\nabla_z\mathcal{D} \bigl(I(x_i), I(x_i + Z)\bigr) \bigr] \mathbb{E}\bigl[\nabla_z\mathcal{D} \bigl(I(x_i), I(x_i + Z)\bigr) \bigr]^\top \biggr].
\end{equation*}
Therefore, for the Euclidean norm constraint $\Vert\cdot\Vert_2$, the optimal solution $\delta^*$ is aligned with the top eigenvector of matrix $M$. However, note that $M$ satisfies the following identity:
\begin{equation*}
    M = \frac{1}{n} GG^\top.
\end{equation*}
As a result, $\delta^*$ follows from the normalized top right singular vector of matrix $G$, and the proof is complete.

\begin{table*}
\begin{center}
\caption{Percentage of fooling rates in VGG-16, MobileNet, and 2-layer CNN trained on Tiny-ImageNet, CIFAR-10, and MNIST.}
\label{table:table2}
\resizebox{0.95\textwidth}{!}{%
\begin{tabular}{llcc|cc|ccc}
\toprule
{} & {} & \multicolumn{2}{c}{\textbf{Tiny-ImageNet}} & \multicolumn{2}{c}{\textbf{CIFAR-10}} & \multicolumn{3}{c}{\textbf{MNIST}} \\\\
Attack & Type & VGG-16 & MobileNet & VGG-16 & MobileNet & MobileNet & 2-layer CNN & 2-layer FCN\\
\midrule
\multirow{2}{*}{Random} & Per image & 1.18 & 3.61 & 0.66 & 1.08 & 0.08 & 0.49 & 0.05 \\
\cmidrule{2-9}
{} & Universal & 1.18 & 3.94 & 0.63 & 1.15 & 0.06 & 0.54 & 0.05 \\
\cmidrule{1-9}
\multirow{4}{*}{Interpretation (Simple)} & Per image-PGD & 76.58 & 92.64 & 34.11 & 30.28 & 5.46 & 0.66 & 0.16 \\
\cmidrule{2-9}
{} & UPI-PCA-PGD (Ours) & 5.59 & 5.03 & 4.73 & 4.12 & 0.24 & 2.05 & 0.39 \\
{} & UPI-PCA-FGM (Ours) & 4.46 & 4.23 & 4.57 & 2.98 & 0.24 & 2.17 & 0.45 \\
{} & UPI-Grad (Ours) & 5.32 & 10.27 & 2.54 & 4.78 & 0.79 & 0.82 & 0.51 \\
\cmidrule{1-9}
\multirow{4}{*}{Interpretation (Integrated)} & Per image-PGD & 60.33 & 47.72 & 25.19 & 36.28 & 5.85 & 6.45 & 2.58 \\
\cmidrule{2-9}
{} & UPI-PCA-PGD (Ours) & 5.93 & 5.77 & 4.62 & 3.38 & 0.26 & 2.17 & 0.67 \\
{} & UPI-PCA-FGM (Ours) & 5.62 & 4.63 & 4.78 & 4.00 & 0.26 & 2.15 & 0.63 \\ 
{} & UPI-Grad (Ours) & 7.41 & 20.48 & 5.61 & 5.62 & 0.77 & 1.10 & 0.76 \\
\bottomrule
\end{tabular}
}
\end{center}
\end{table*}

\section{Additional Numerical Results}
\begin{figure*}[h!]
    \centering
    \includegraphics[width=0.9\textwidth]{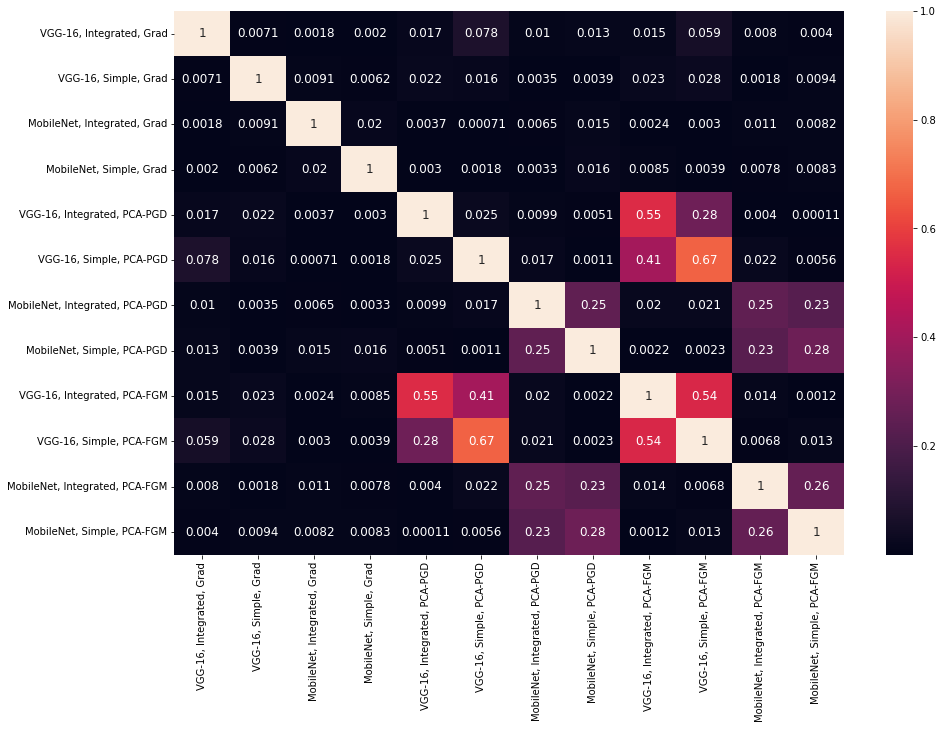}
    \caption{Cross-correlation between generated UPIs in the Tiny-ImageNet experiments.}
    \label{Figure:Correlation_Tiny}
\end{figure*}

\begin{figure*}[h!]
    \centering
    \includegraphics[width=0.9\textwidth]{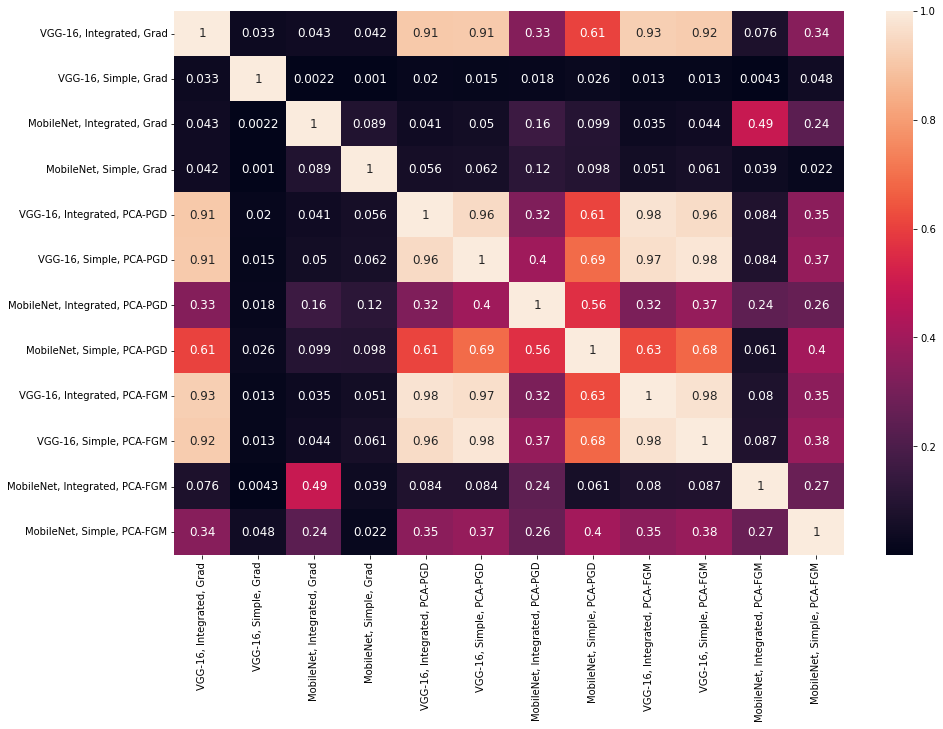}
    \caption{Cross-correlation between generated UPIs in the CIFAR-10 experiments.}
    \label{Figure:Correlation_CIFAR10}
\end{figure*}

\begin{figure*}[h!]
    \centering
    \includegraphics[width=0.9\textwidth]{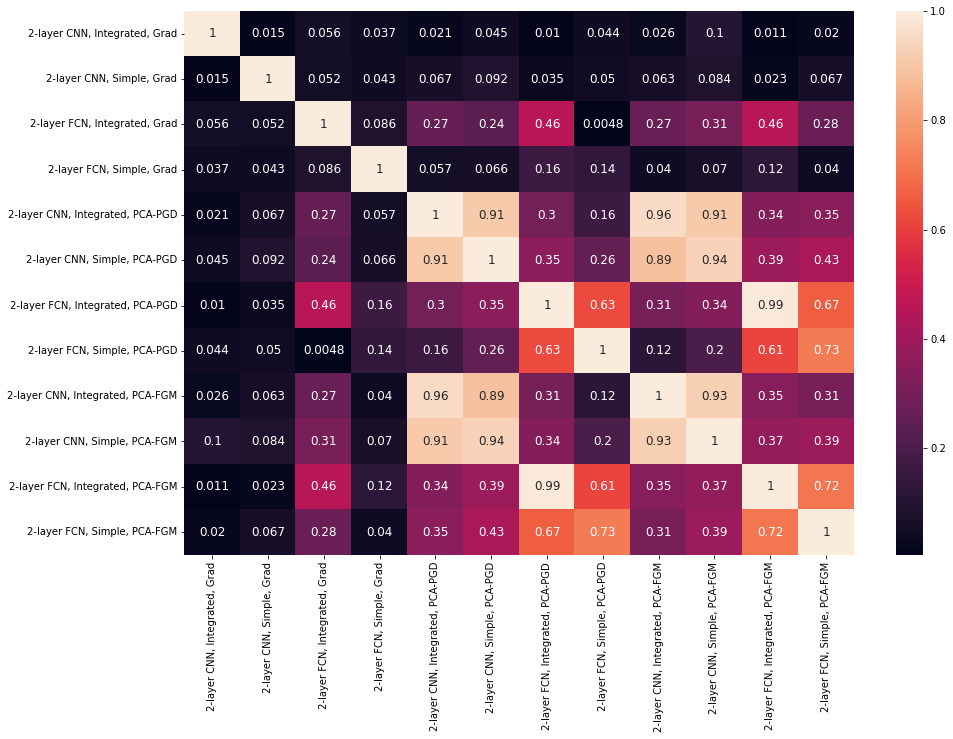}
    \caption{Cross-correlation between generated UPIs in the MNIST experiments.}
    \label{Figure:Correlation_MNIST}
\end{figure*}

\begin{figure*}[h!]
    \centering
    \small
    \stackunder[0pt]{
    \stackunder[0pt]{\includegraphics[width=0.2\textwidth]{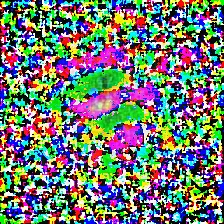}}{}
    \stackunder[0pt]{\includegraphics[width=0.2\textwidth]{images/tiny/vgg16_simple_fgm.jpg}}{}
    \stackunder[0pt]{\includegraphics[width=0.2\textwidth]{images/tiny/vgg16_simple_grad.jpg}}{}
    }{}
    \stackunder[5pt]{
    \stackunder[5pt]{\includegraphics[width=0.2\textwidth]{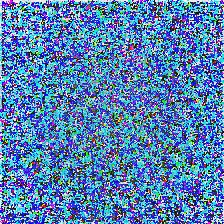}}{UPI-PCA-PGD}
    \stackunder[5pt]{\includegraphics[width=0.2\textwidth]{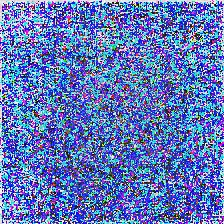}}{UPI-PCA-FGM}
    \stackunder[5pt]{\includegraphics[width=0.2\textwidth]{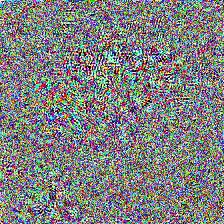}}{UPI-Grad}
    }{(a) Simple Gradient}
    \\
    \vspace{10pt}
    \stackunder[0pt]{
    \stackunder[0pt]{\includegraphics[width=0.2\textwidth]{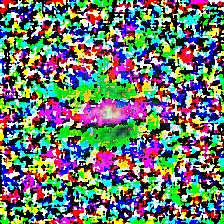}}{}
    \stackunder[0pt]{\includegraphics[width=0.2\textwidth]{images/tiny/vgg16_integrated_fgm.jpg}}{}
    \stackunder[0pt]{\includegraphics[width=0.2\textwidth]{images/tiny/vgg16_integrated_grad.jpg}}{}
    }{}
    \stackunder[5pt]{
    \stackunder[5pt]{\includegraphics[width=0.2\textwidth]{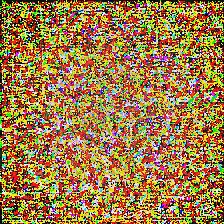}}{UPI-PCA-PGD}
    \stackunder[5pt]{\includegraphics[width=0.2\textwidth]{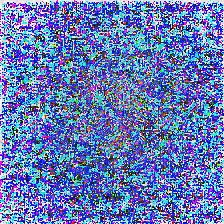}}{UPI-PCA-FGM}
    \stackunder[5pt]{\includegraphics[width=0.2\textwidth]{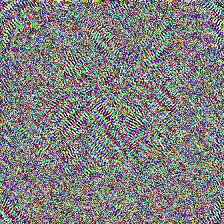}}{UPI-Grad}
    }{(b) Integrated Gradients}
    \\
    \vspace{10pt}
    \stackunder[5pt]{
    \stackunder[5pt]{\includegraphics[width=0.2\textwidth]{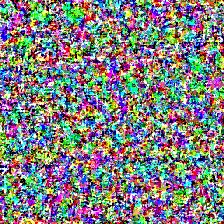}}{VGG-16}
    \stackunder[5pt]{\includegraphics[width=0.2\textwidth]{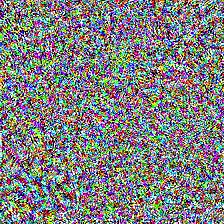}}{MobileNet}
    }{(c) UAP}
    \caption{Univesal perturbations for Tiny-ImageNet. The top and bottom rows of groups (a) and (b) are optimized for VGG-16 and MobileNet, respectively.}
    \label{Figure:UPI_Tiny}
\end{figure*}

\begin{figure*}[h!]
    \centering
    \small
    \stackunder[0pt]{
    \stackunder[0pt]{\includegraphics[width=0.2\textwidth]{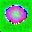}}{}
    \stackunder[0pt]{\includegraphics[width=0.2\textwidth]{images/cifar10/vgg16_simple_fgm.jpg}}{}
    \stackunder[0pt]{\includegraphics[width=0.2\textwidth]{images/cifar10/vgg16_simple_grad.jpg}}{}
    }{}
    
    \stackunder[5pt]{
    \stackunder[5pt]{\includegraphics[width=0.2\textwidth]{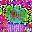}}{UPI-PCA-PGD}
    \stackunder[5pt]{\includegraphics[width=0.2\textwidth]{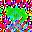}}{UPI-PCA-FGM}
    \stackunder[5pt]{\includegraphics[width=0.2\textwidth]{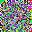}}{UPI-Grad}
    }{(a) Simple Gradient}
    \\
    \vspace{10pt}
    \stackunder[0pt]{
    \stackunder[0pt]{\includegraphics[width=0.2\textwidth]{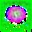}}{}
    \stackunder[0pt]{\includegraphics[width=0.2\textwidth]{images/cifar10/vgg16_integrated_fgm.jpg}}{}
    \stackunder[0pt]{\includegraphics[width=0.2\textwidth]{images/cifar10/vgg16_integrated_grad.jpg}}{}
    }{}
    \stackunder[5pt]{
    \stackunder[5pt]{\includegraphics[width=0.2\textwidth]{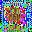}}{UPI-PCA-PGD}
    \stackunder[5pt]{\includegraphics[width=0.2\textwidth]{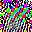}}{UPI-PCA-FGM}
    \stackunder[5pt]{\includegraphics[width=0.2\textwidth]{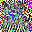}}{UPI-Grad}
    }{(b) Integrated Gradients}
    \\
    \vspace{10pt}
    \stackunder[5pt]{
    \stackunder[5pt]{\includegraphics[width=0.2\textwidth]{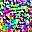}}{VGG-16}
    \stackunder[5pt]{\includegraphics[width=0.2\textwidth]{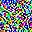}}{MobileNet}
    }{(c) UAP}
    \caption{Universal perturbations for CIFAR-10 data. The top and bottom rows of groups (a) and (b) are optimized for VGG-16 and MobileNet, respectively.}
    \label{Figure:UPI_CIFAR10}
\end{figure*}

\begin{figure*}[h!]
    \centering
    \small
    \stackunder[0pt]{
    \stackunder[0pt]{\includegraphics[width=0.2\textwidth]{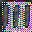}}{}
    \stackunder[0pt]{\includegraphics[width=0.2\textwidth]{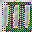}}{}
    \stackunder[0pt]{\includegraphics[width=0.2\textwidth]{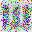}}{}
    }{}
    \stackunder[0pt]{
    \stackunder[0pt]{\includegraphics[width=0.2\textwidth]{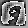}}{}
    \stackunder[0pt]{\includegraphics[width=0.2\textwidth]{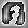}}{}
    \stackunder[0pt]{\includegraphics[width=0.2\textwidth]{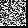}}{}
    }{}
    \stackunder[5pt]{
    \stackunder[5pt]{\includegraphics[width=0.2\textwidth]{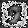}}{UPI-PCA-PGD}
    \stackunder[5pt]{\includegraphics[width=0.2\textwidth]{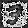}}{UPI-PCA-FGM}
    \stackunder[5pt]{\includegraphics[width=0.2\textwidth]{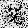}}{UPI-Grad}
    }{}
    \caption{UPIs using Simple Gradient on MNIST. The top, middle, and bottom rows are found for MobileNet, 2-layer CNN, and 2-layer FCN respectively.}
    \label{Figure:UPI_MNIST_Simple}
\end{figure*}

\begin{figure*}[h!]
    \centering
    \small
    \stackunder[0pt]{
    \stackunder[0pt]{\includegraphics[width=0.2\textwidth]{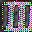}}{}
    \stackunder[0pt]{\includegraphics[width=0.2\textwidth]{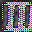}}{}
    \stackunder[0pt]{\includegraphics[width=0.2\textwidth]{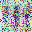}}{}
    }{}
    \\
    \stackunder[0pt]{
    \stackunder[0pt]{\includegraphics[width=0.2\textwidth]{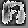}}{}
    \stackunder[0pt]{\includegraphics[width=0.2\textwidth]{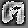}}{}
    \stackunder[0pt]{\includegraphics[width=0.2\textwidth]{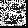}}{}
    }{}
    \stackunder[5pt]{
    \stackunder[5pt]{\includegraphics[width=0.2\textwidth]{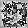}}{UPI-PCA-PGD}
    \stackunder[5pt]{\includegraphics[width=0.2\textwidth]{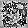}}{UPI-PCA-FGM}
    \stackunder[5pt]{\includegraphics[width=0.2\textwidth]{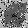}}{UPI-Grad}
    }{}
\caption{UPIs designed for Integrated Gradients on MNIST data. The top, middle, and bottom rows are optimized for MobileNet, 2-layer CNN, and 2-layer FCN respectively.}
\label{Figure:UPI_MNIST_Integrated}
\end{figure*}

\begin{figure*}[h!]
    \centering
    \small
    \stackunder[5pt]{\includegraphics[width=0.2\textwidth]{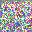}}{MobileNet}
    \stackunder[5pt]{\includegraphics[width=0.2\textwidth]{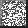}}{2-layer CNN}
    \stackunder[5pt]{\includegraphics[width=0.2\textwidth]{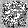}}{2-layer FCN}
\caption{UAPs optimized for the MNIST data.}
\label{Figure:UAP_MNIST}
\end{figure*}

While increasing the value of fooling rates is not our goal in this paper, they are presented in Table \ref{table:table2}. Our results demonstrate that even though most UPIs cause a lower fooling rate than per-image perturbations, they have a considerable impact on the interpretations. 
We further studied the transferability of UPIs on the datasets mentioned above. As shown in Figure~\ref{Figure:Correlation_Tiny}, \ref{Figure:Correlation_CIFAR10}, \ref{Figure:Correlation_MNIST}, UPI-PCA method has superior generalizability over UPI-Grad. Moreover, UPIs that are generated using Integrated Gradients are strongly correlated with their Simple Gradient counterparts. Additionally, our numerical results in Figure~\ref{Figure:UPI_Tiny}, \ref{Figure:UPI_CIFAR10}, \ref{Figure:UPI_MNIST_Simple}, \ref{Figure:UPI_MNIST_Integrated}, \ref{Figure:UAP_MNIST} suggest that UPIs are visually structured. UAPs are similarly illustrated for a better comparison. Notably, some of the universal adversarial perturbations found for MNIST have number-like representations.

